\pdfoutput=1
\documentclass[11pt]{article}

\usepackage[]{acl}

\usepackage{amsmath,amsfonts}
\usepackage{algorithmic}
\usepackage{array}
\usepackage[caption=false,font=normalsize,labelfont=sf,textfont=sf]{subfig}
\usepackage{textcomp}
\usepackage{stfloats}
\usepackage{url}
\usepackage{verbatim}
\usepackage{graphicx}

\usepackage{multirow}
\usepackage{enumitem}
\usepackage{amsmath}
\usepackage{amssymb}
\usepackage{ulem}   % underline
\usepackage{subcaption}
\usepackage{makecell}
\usepackage{color}
\usepackage{booktabs}
\usepackage{times}
\usepackage{latexsym}
\usepackage{float}

\usepackage[T1]{fontenc}
\usepackage[utf8]{inputenc}

\usepackage{microtype}

\usepackage{inconsolata}

\title{CADGE: Context-Aware Dialogue Generation Enhanced with Graph-Structured Knowledge Aggregation}

\author{Chen Tang\textsuperscript{1}, Hongbo Zhang\textsuperscript{2}, Tyler Loakman\textsuperscript{2},  Bohao Yang\textsuperscript{1}, \\
\textbf{Stefan Goetze\textsuperscript{2}},\textbf{ Chenghua Lin\textsuperscript{1}\thanks{corresponding author.}}\\
    \textsuperscript{1}Department of Computer Science, The University of Manchester, UK \\
    \textsuperscript{2}Department of Computer Science, The University of Sheffield, UK \\
  \texttt{\{chen.tang,chenghua.lin\}@manchester.ac.uk} \\ 
  \texttt{bohao.yang-2@postgrad.manchester.ac.uk} \\ 
  \texttt{\{hzhang183,tcloakman1,s.goetze\}@sheffield.ac.uk}}
         
\graphicspath{{figs/}}

\begin{document}
\maketitle
\begin{abstract}
Commonsense knowledge is crucial to many natural language processing tasks. Existing works usually incorporate graph knowledge with conventional graph neural networks (GNNs), resulting in a sequential pipeline that compartmentalizes the encoding processes for textual and graph-based knowledge. This compartmentalization does, however, not fully exploit the contextual interplay between these two types of input knowledge. In this paper, a novel context-aware graph-attention model (Context-aware GAT) is proposed, designed to effectively assimilate global features from relevant knowledge graphs through a context-enhanced knowledge aggregation mechanism. Specifically, the proposed framework employs an innovative approach to representation learning that harmonizes heterogeneous features by amalgamating flattened graph knowledge with text data. The hierarchical application of graph knowledge aggregation within connected subgraphs, complemented by contextual information, to bolster the generation of commonsense-driven dialogues is analyzed. Empirical results demonstrate that our framework outperforms conventional GNN-based language models in terms of performance. Both, automated and human evaluations affirm the significant performance enhancements achieved by our proposed model over the concept flow baseline.
\end{abstract}

% ----------- fig:intro -----------
\begin{figure}[t]
\centering
\includegraphics[width=0.9\columnwidth]{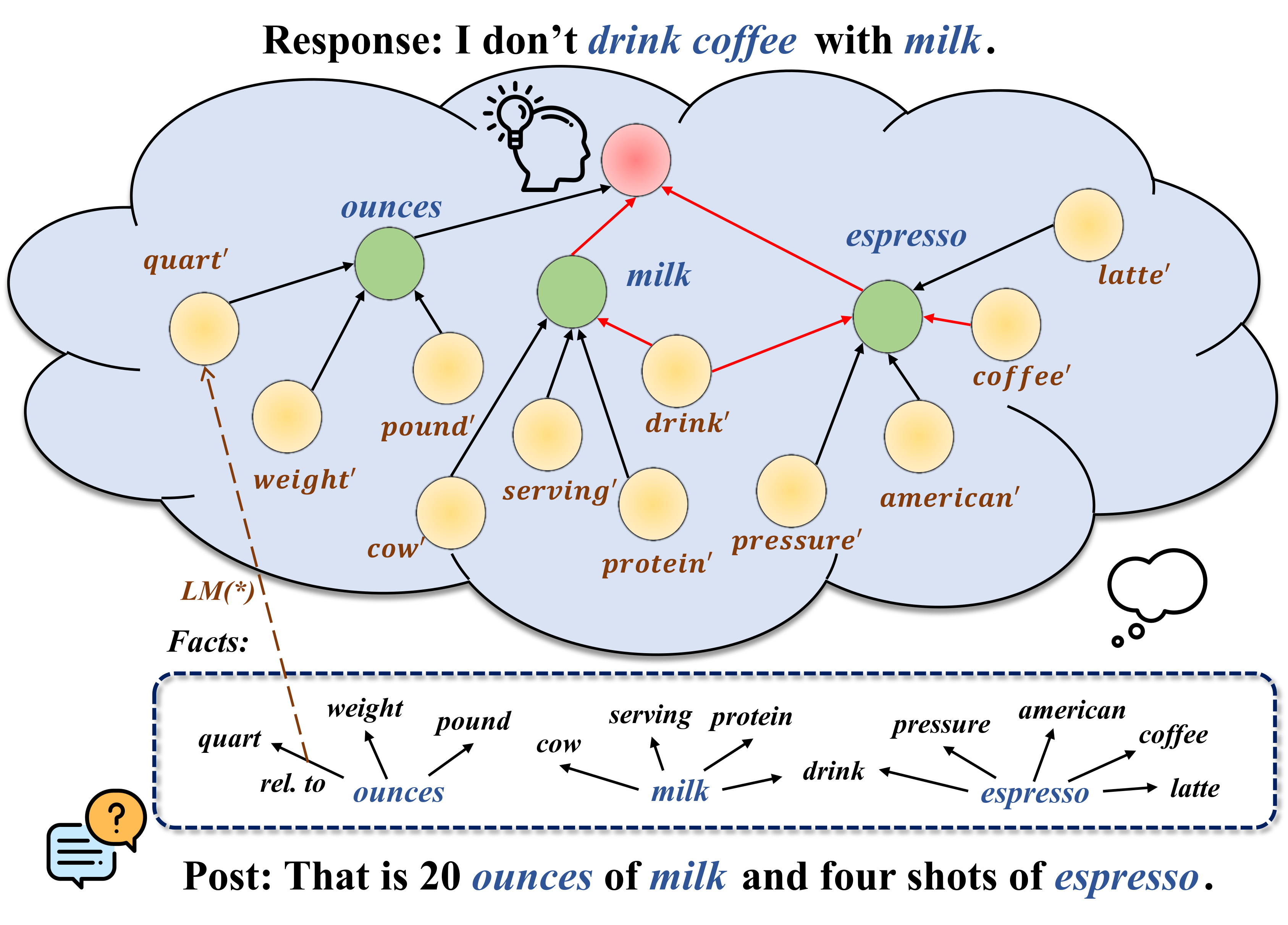}
\caption{Illustration of the proposed model with an example. The retrieved facts are fed to the graph model, then the model learns the representations of concepts by aggregating the knowledge layer by layer. Finally, responses are generated with these aggregated features.} 
\label{fig:intro}
\end{figure}
% ----------- end of fig -----------

% =============================== Section 1 ==================================
\section{Introduction}
% briefly introduce the background
Open-domain dialogue generation has gained considerable traction in the field of natural language generation \cite{roller-etal-2021-recipes,tang-etal-2023-enhancing}. This task aims to develop chatbots with the capacity to engage in conversations across a broad spectrum of topics, thereby enabling a multitude of practical applications, including virtual assistants and well-being support systems~\cite{abd-yusof-etal-2017-analysing,wang-etal-2021-fast,tang2023terminology,yang2024improving}.
In recent years, there has been a surge of interest in leveraging large language models for dialogue generation \cite{diagpt,meena,roller-etal-2021-recipes,tang-etal-2022-ngep,huang-etal-2022-improving}. These models, in general, exhibit an enhanced capacity to encapsulate knowledge within their networks as their model sizes increase. However, it is crucial to acknowledge a series of studies that have underscored the limitations of training on plain text corpora, where the knowledge structure is not explicitly represented during the learning process~\cite{tang-etal-2022-etrica,yang-etal-2024-effective-distillation}. Consequently, a key research question concerns how to better exploit and use external knowledge to improve the quality of generated responses, which has received increasing attention in recent research~\cite{zhang-etal-2020-grounded,yu-etal-2022-diversifying-content,wu2022improving,tang-etal-2023-improving}.

% Introduce existing works and summarize their difficulty and drawbacks
The knowledge incorporated into chatbots can be broadly divided into \textit{structured} and \textit{unstructured} forms. Prior work \cite{komeili-etal-2022-internet,ghazvininejad2018knowledge,lian2019response} has achieved successful integration of unstructured knowledge (such as free-text content from web pages and knowledge statements) into the generated responses of chatbots. This typically involves encoding the most appropriate retrieved facts together with the conversation context or encoding multiple pieces of facts into a uniform representation before passing it to the decoder alongside the conversation history. Structured knowledge, on the other hand, usually takes the form of a graph. A range of neural models \cite{zhou2018commonsense, Yang2020GraphDialogIG, Lin2021KnowledgeAwareGG} have been introduced to incorporate features from the retrieved graph-structured knowledge. For instance, the graph attention mechanism \cite{lotfi-etal-2021-teach, tuan2019dykgchat, zhou2018commonsense} has been widely used to embed knowledge graph features, and has been successful in aggregating sparse features into rich representations. With regard to language models, the rise of pre-trained models \cite{srivastava2021pretrain, dong2019unified,tang2024cross} has also substantially advanced the state-of-the-art (SOTA) in open-domain dialogue generation.

However, existing dialogue systems still face a number of challenges to effectively exploit commonsense knowledge \cite{xie2021survey}. Since graph-structured knowledge and natural utterances have different representations, most prior work \cite{tuan2019dykgchat, zhou2018commonsense,zhang-etal-2020-grounded} employed separate encoders to incorporate and leverage these heterogeneous features by concatenating their respective numeric vectors. However, since the separate encoders do not share low-dimensional representations, they may fail to fully account for the semantics of context contributed by given posts with additional external knowledge facts. In addition, existing frameworks directly conduct graph-attention-based encoding on retrieved facts from the knowledge base, which are isolated in separate sub-graphs. This strategy does not capture dependencies between sub-graphs nor between the graph knowledge and the context of the post, in turn making it hard for neural networks to fully capture the overall backgrounds from the inputs.

% Introduce our work 
To address the aforementioned challenges, this paper proposes a novel graph-based framework to leverage knowledge contained in concept-related facts. In contrast to employing separate encoders to encode knowledge in the form of disparate knowledge graphs and text, we first transform the graph-structured representations into plain text, and leverage a pre-trained language model, UniLM \cite{dong2019unified}, to generate unified features for all inputs. Subsequently, to overcome inadequacies when capturing the context semantics provided by the given posts and retrieved knowledge facts, a novel, context-aware graph-based mechanism (Context-aware GAT) is proposed to incorporate the features from the post and the knowledge graph in the same learning process during hierarchical aggregation. The graph knowledge takes two steps (layers) before being aggregated into a condensed feature vector as the global features of given inputs. For each layer, the context embedding and the factual embedding are concatenated, and then graph attentions are computed for every sub-node. Finally, all representations are aggregated into the root node and fed to the decoder for response generation. This whole process is illustrated in \autoref{fig:intro}. We also note that our model can be easily extended to incorporate multi-hop knowledge. Experimental results show that our extended model can use multi-hop knowledge to further increase the informativeness of generated responses, and consequently yields considerable improvements over other dialogue systems that use multi-hop knowledge.
% Mention our contributions
The contributions of this work are summarised three-fold:
\begin{itemize}
[noitemsep,nolistsep,leftmargin=*]
\item We propose a novel framework\footnote{ Our code and datasets are accessible at \url{https://github.com/StevenZHB/CADGE}.}, which is a successful exploration that leverages a unified language model for the heterogeneous inputs of graph knowledge and text, exploiting structured knowledge with context-aware subgraph aggregation to generate informative responses. 
\item We conduct a range of experiments, and the extensive automatic and human evaluation results demonstrate our model significantly outperforms existing baselines to generate a more appropriate and informative response with external graph knowledge. 
\item With extensive experiments, we investigate the advances and mechanisms of leveraging graph knowledge with our Context-aware GAT model. We also investigate the expansion of our model to accommodate multi-hop knowledge, and validate its effectiveness.
\end{itemize}

% =============================== Section 2 ==================================
\section{Related Work}

%\vspace{-3mm}

Recently, much work has focused on augmenting dialogue systems with additional background knowledge. Such works can be divided into dialogue systems augmented with unstructured knowledge, and those augmented with structured knowledge. With unstructured knowledge, \cite{komeili-etal-2022-internet} models web page information and feeds it into a language model. \cite{ghazvininejad2018knowledge} and \cite{lotfi-etal-2021-teach} encode the filtered factual statements with a specific encoder and then pass them into the decoder along with context. \cite{lian2019response} use context to aggregate knowledge statements and find that aggregated knowledge gives better results than filtered knowledge. Regarding structured knowledge, graph neural networks \cite{GNN} are usually used to embed graph information to input into a language model. \cite{zhou2018commonsense} uses GRUs and two graph attention modules to select appropriate triples to incorporate into responses. In order to exploit the benefits of multi-hop knowledge, \cite{zhang-etal-2020-grounded} adds an attention mechanism in a similar way to filter the appropriate knowledge. Finally, \cite{tuan2019dykgchat} proposes a model which selects the output from a sequence-to-sequence model and a multi-hop reasoning model at each time step.

%\subsection{Open-domain Dialogue Generation}
Large language models such as UniLM \cite{dong2019unified}, GPT-2 \cite{radford2019language}, and BART \cite{bart} are widely used in open domain dialogue generation systems~\cite{zeng-etal-2021-affective}. DialoGPT \cite{diagpt} was pre-trained on a dialogue dataset containing 147M conversations and is based on the autoregressive GPT-2 model, using a maximum mutual information (MMI) scoring function to address the low amount of information in the generated text. \cite{meena} built a 2.6B-parameter Evolved Transformer architecture to model the relation between context-response pairs. To generate more informative responses, \cite{bao2019plato,bao2020plato} use latent variables to model one-to-many relationships in context-response pairs. Finally, \cite{roller-etal-2021-recipes} use a retrieval model to retrieve candidate responses and then concatenates them to represent the context before inputting them into the transformer to generate the model. Please refer to Appendix \ref{apx:related-works} for more details of related work.

% =============================== Section 3 ==================================
\section{Methodology}
\label{sec:methodology}

\begin{figure}[t]
\centering
\includegraphics[width=0.8\columnwidth]{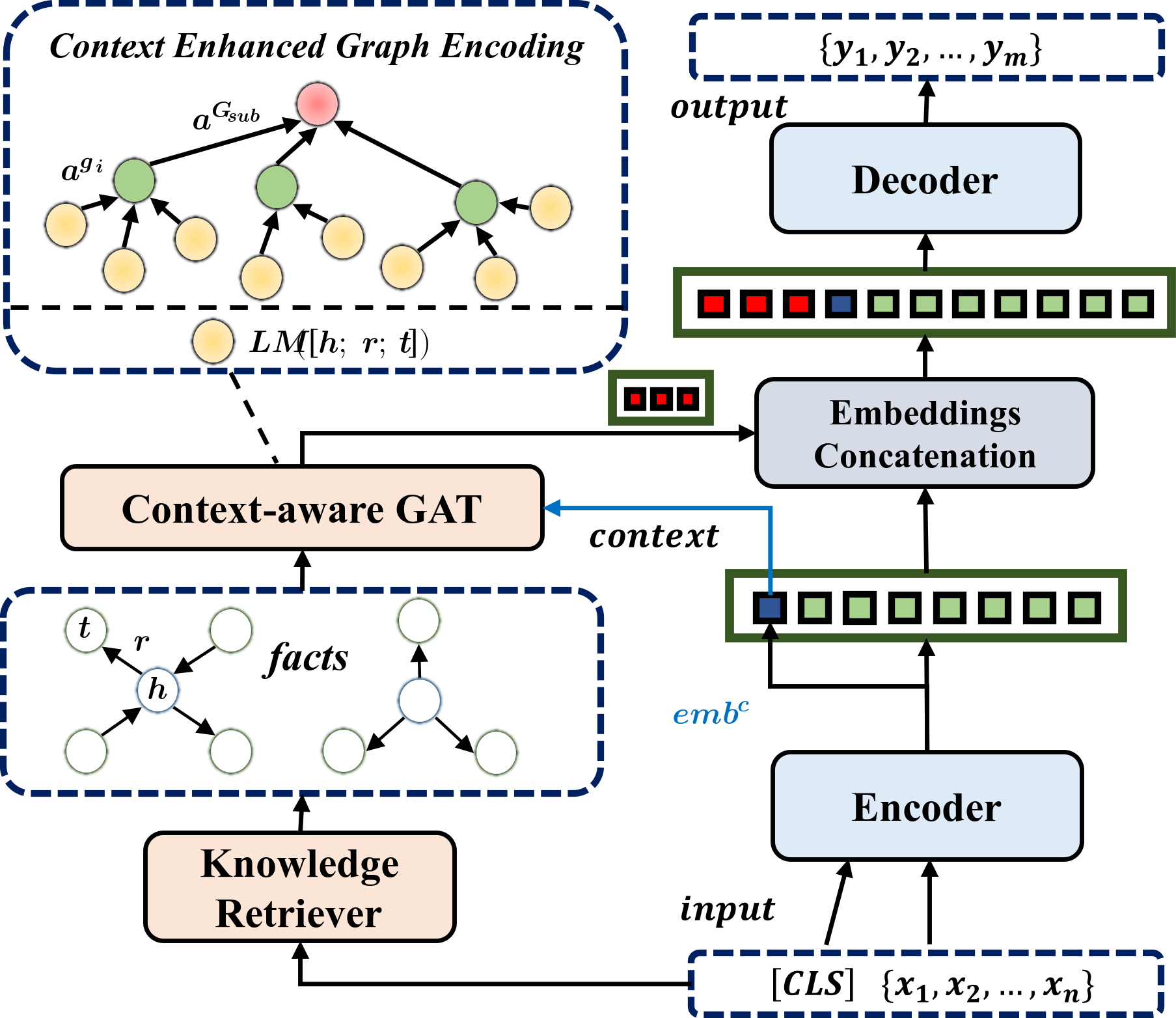}
\caption{Overview of the proposed model. } 
\label{fig:overview}
\end{figure}

We formulate our task as follows: The given inputs include a post $ X = \{x_1, x_2, ..., x_n\} $ and a graph knowledge base $ G = \{\tau_1, \tau_2, ..., \tau_k\}$, in which a fact is represented in the form of a triplet $ \{h, r, t\}$ where $h$, $r$, and $t$ denote the head node, the relation, and the tail node, respectively. The goal is to generate a response $ Y = \{y_1, y_2, ..., y_m\} $ by modeling the conditional probability distribution $ P(Y|X, G)$. \autoref{fig:overview} gives an overview of our framework. The knowledge retrieval process is fundamentally implemented by word matching (concepts in ConceptNet are formatted in one-word) and rule filtering to collect knowledge triples (for more details please refer to \cite{zhou2018commonsense}).

% ----------- fig:overview -----------
% ----------- end of fig -----------

% ----------- fig:gat -----------
\begin{figure}[t]
\centering
\includegraphics[width=0.95\columnwidth]{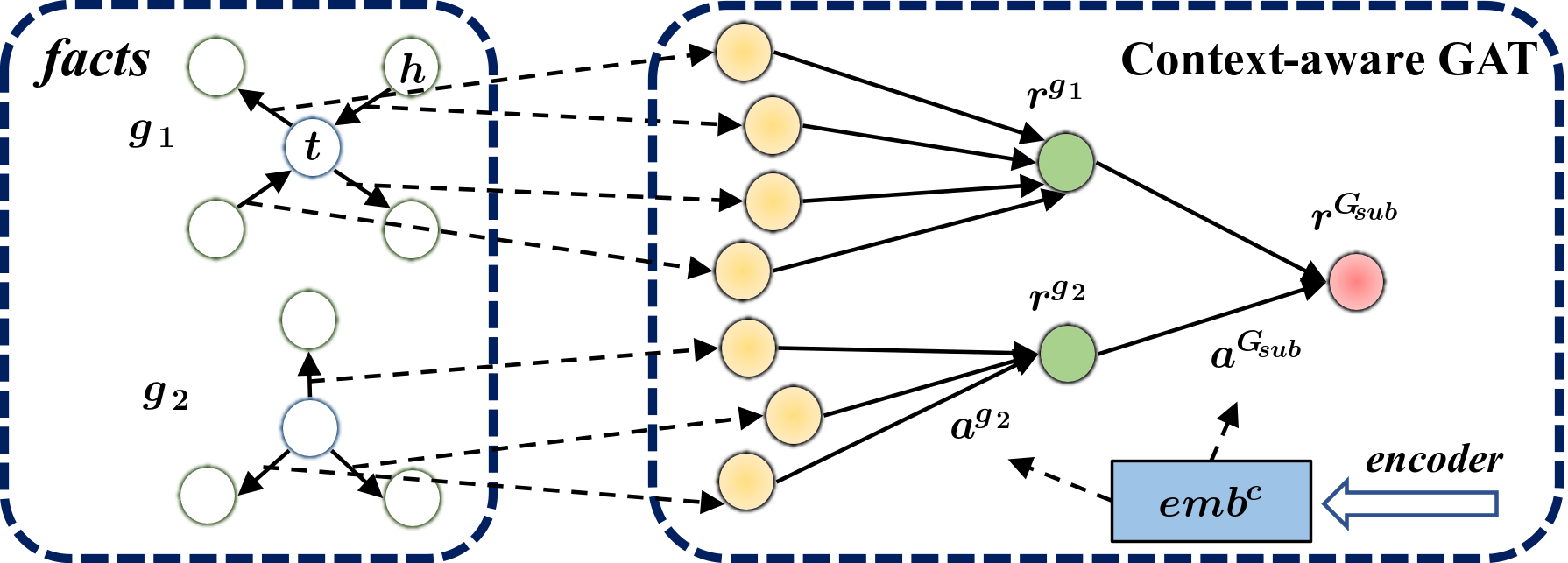}
\caption{The Context-aware GAT firstly transforms knowledge from facts into numeric vectors (in yellow). Through feature forwarding, the root nodes of each graph attentively read and aggregate all knowledge and become higher-level representations (from yellow to green, and then green to red).}
\label{fig:gat}
\end{figure}
% ----------- end of fig -----------

% ######################### Section 3.1 ########################
\subsection{Knowledge Representation}
\label{sec:knowledgerepresentation}
The 12-layer transformer blocks of UniLM \cite{dong2019unified} are split into two 6-layer parts - the encoder and decoder. When encoding the post's text, the language model of UniLM is informed of the high-level narrative structure using a classification label ([CLS]) to allow learning of the overall representation from $ X $ as the context feature $emb^{c}$. For each recognised entity $ent_i$ in the post, relevant facts are retrieved from the knowledge base in the form of triples, and all retrieved facts can be considered as sub-graphs $ g_{i} = \{\tau_1, \tau_2, ..., \tau_{N_{g_i}}\} $ in $ G $. Each post usually results in several independent sub-graphs $G_{sub} = \{g_1, g_2, ..., g_{N_{G_{sub}}}\} $. 
In contrast to existing works that encode knowledge in the form of disparate knowledge graphs and text, 
we propose to transform facts into text by directly concatenating them into a string, where they are then encoded with the embedding layer of UniLM:
\begin{align}
    E^{post} &= LM([l_{[CLS]}; \{x_1, ...\}])) \label{eq:E-post} \\
    & = \{emb^c, emb_1 ...\} \\
    f_e({h, r, t}) & = LM_{emb}([h; r; t]) \label{eq:f-e} \\
    E^{\tau} & = f_e({h, r, t})\quad \mathrm{s.t.} \{h, r, t\} \in g_i
\end{align}
Operator $LM$ (abbr. of language model) denotes the encoder of  UniLM, whilst $LM_{emb}$ denotes the embedding layer of UniLM, and $l_{[CLS]}$ denotes the ``[CLS]'' label.

% ######################### Section 3.2 ########################
\subsection{Context-aware GAT}
\label{sec:contextawaregat}
The overview of the proposed Context-aware GAT is as illustrated in \autoref{fig:gat}. The model learns the global graph features via translations operating on both the low-dimensional embeddings of the knowledge facts and the context contained in $emb^{c}$.
To facilitate knowledge understanding and generation, we leverage a graph attention mechanism to aggregate knowledge representations layer by layer. With two layers of feature forward processing, we obtain the representation of the root node, $rt_{G_{sub}}$, as the aggregated feature for the whole graph, $G_{sub}$.

%\subsubsection{First Forward Layer}
\noindent\textbf{First Forward Layer.}
Our model firstly attends to the representations of facts $\tau \in g_i$ to compute graph attention and then aggregates features to the root node of each graph $rt^{g_i}$. The knowledge gradually updates the representations of root nodes step by step:
\begin {align}
    % rt_{0}^{g_i} & = f_e({l_{pad},l_{pad},l_{pad}})  \\
    rt_{t}^{g_i} & = \sum_{j=1}^{N_{g_i}} a_{tj}^{g_i} E_{tj}^{\tau} \\
    a_{tj}^{g_i} & = \frac{\exp(\beta_{tj}^{g_i})}{\sum_{j=1}^{N_{g_i}} \exp(\beta_{tj}^{g_i})} \\
    \beta_j^{g_i} & = W^{g_i}{[E_{tj}^{\tau}; emb^c]}^{\mathrm{T}}
\end{align}
where $t$ denotes the time step, $l_{pad}$ denotes the padding label to help initialize the root representations, and $W^{g_i}$ is a trainable parameter matrix.

%\subsubsection{Second Forward Layer}
\noindent\textbf{Second Forward Layer.}
In analogy to the first forward layer, our model attends to the root nodes $rt^{g_i}$ represented for each sub-graph to attentively compute the final representation of the root node $rt^{G_{sub}}$, which stands for the overall features of all the retrieved sub-graphs:
\begin{align}
    % rt_{0}^{G_{sub}} & = f_e({l_{pad},l_{pad},l_{pad}})  \\
    rt_{t}^{G_{sub}} & = \sum_{i=1}^{N_{G_{sub}}} a_{ti}^{G_{sub}} (rt_{t}^{g_i}) \\
    a_{ti}^{G_{sub}} & = \frac{\exp(\beta_{ti}^{G_{sub}})}{\sum_{i=1}^{N_{G_{sub}}} \exp(\beta_{tj}^{G_{sub}})} \\
    \beta_i^{G_{sub}} & = W^{G_{sub}}{[rt_{t}^{g_i}; emb^c]}^{\mathrm{T}}
    % rt_0^{g_i} &= E^{\tau}_{i}
\end{align}

% ######################### Section 3.3 ########################
\subsection{Feature Aggregation and Decoding}
\label{sec:featureaggregation}
After computing a representation for the root node, features from the post and retrieved knowledge are concatenated, and the decoder is employed to predict tokens $Y$ as the output response:
% ----------- equation-----------
\begin{align}
    & V = [rt^{G_{sub}}; E^{post}] \label{eq: one-to-decoder} \\
    & H = \text{Decoder}(V) \\
    & P(Y|X) = \text{softmax}(H W)
\end{align}
% ----------- end of equation -----------
where $V$ denotes the aggregated features fed to the decoder, $H$ denotes the hidden states of the decoder used to predict the probability distribution of output tokens $P(Y|X)$, and $W$ is a trainable parameter.

% ######################### Section 3.4 ########################
\subsection{Loss Function}
%\subsubsection{Auxiliary Entity Selection Task}
\noindent\textbf{Auxiliary Entity Selection Task.} 
To better support representation learning, the entity selection task is introduced as an auxiliary task when training the proposed generative system. For each input post, the datasets contain corresponding annotations of knowledge triples $ \Gamma=\{\tau_1^{'}, ..., \tau_{N_{\Gamma}}^{'}\}$ from the knowledge base. These annotations can be considered as the ground truth of the knowledge paired with the post. The neural model is forced to select the ground-truth triples from all retrieved knowledge $G_{sub} $. As \autoref{fig:gat} shows, each yellow node represents a knowledge triplet $\tau$, and each green node represents the root node $rt^{g_i}$. All yellow and green nodes have been labeled by checking if they are annotated as the ground truth. For instance, if $\tau_j \in \Gamma$ then the probability of $\tau_j$ denoted as $p_{es}(\tau_j|X)$ should be $1$, and $0$ otherwise. For the sub-graph root node (the green node in \autoref{fig:gat}), if $\tau_j \in g_i$ is the truth, then $p_{es}(rt^{g_i}|X)$ should be 1, and 0 otherwise. The probability distribution is modelled as follows:
% ----------- equation-----------
\begin{align}
    p_{es}(\tau_j|X) & = \text{softmax}(E_{j}^{\tau} W^{p_{\tau}}) \\
    p_{es}(rt^{g_i}|X) & = \text{softmax}((rt^{g_i})W^{p_{g_i}})
\end{align}
% ----------- end of equation -----------
where $es$ denotes entity selection, and $W$ denotes the trainable parameters.

% ######################### Section 3.5 ########################
%\subsubsection{Overall Loss Function}
\noindent\textbf{Overall Loss Function.} 
The loss function includes parts of the text prediction task and entity selection task, and is computed with cross entropy:
% ----------- equation-----------
\begin{align}
    \mathcal{L}_{lm} & = - \frac{1}{N}\sum_{n=1}^N \log P(Y|X) \\
    \mathcal{L}_{es}^{\tau} & =  - \sum_{n=1}^{N}\sum_{j=1}^{N_{\tau}} s_j^{\tau}\log (p_{es}(\tau_j|X)) \\
    \mathcal{L}_{es}^{g} & =  - \sum_{n=1}^{N}\sum_{i=1}^{N_{G_{sub}}} s_g^{\tau}\log (p_{es}(rt^{g_i}|X)) \\
    \mathcal{L}_{overall} & = \mathcal{L}_{lm} + \lambda_1 \mathcal{L}_{es}^{\tau} + \lambda_2 \mathcal{L}_{es}^{g}
\end{align}
% ----------- end of equation -----------
where $N$ denotes the total amount of test data. $\lambda_1$ and $\lambda_2$ denotes the scale factors. $\mathcal{L}_{es}^{\tau}$ and $\mathcal{L}_{es}^{g}$ denote the loss of entity selections on the root nodes for facts $\tau$ and $g_i$, respectively. $\lambda_1$ and $\lambda_2$ are set to $1$ in the following experiments.

\subsection{Expansion for Multi-hop Knowledge}
\label{sec:multi-hop}
We also consider extending our model to incorporate multi-hop knowledge, which might give a further performance boost. Specifically, we extract the two-hop knowledge for all one-hop entities and use the same method to build a graph of two-hop knowledge. 
%The full detail of the expansion process is given in Appendix \ref{apx:expansion}.
%\section{Expansion for Multi-hop Knowledge} \label{apx:expansion}
As the aggregation of two-hop knowledge needs to be related to one-hop knowledge, we use the one-hop knowledge aggregation representation $rt_{G^{one}_{sub}}$ in addition to the "[CLS]" label when aggregating two-hop knowledge. After passing through two layers of GAT, the root node of the two-hop knowledge graph ($G^{two}_{sub}$), $rt_{G^{two}_{sub}}$,  which is treated as the aggregated features of the two-hop knowledge graph, is then concatenated with $rt_{G^{one}_{sub}}$ and input to the Decoder. The attention in the context-aware GAT for the two-hop knowledge graph is as follows:
\begin{align}
    a_{ti}^{G^{two}_{sub}} & = \frac{\exp(\beta_{ti}^{G^{two}_{sub}})}{\sum_{i=1}^{N_{G^{two}_{sub}}} \exp(\beta_{tj}^{G^{two}_{sub}})} \\
    \beta_i^{G^{two}_{sub}} & = W^{G^{two}_{sub}}{[rt_{t}^{g^{two}_i}; rt_{G^{one}_{sub}};emb^c]}^{\mathrm{T}}\\
    a_{tj}^{g^{two}_i} & = \frac{\exp(\beta_{tj}^{g^{two}_i})}{\sum_{j=1}^{N_{g^{two}_i}} \exp(\beta_{tj}^{g^{two}_i})} \\
    \beta_j^{g^{two}_i} & = W^{g^{two}_i}{[E_{tj}^{\tau}; rt_{G^{one}_{sub}};emb^c]}^{\mathrm{T}}
\end{align}
The aggregated feature for the decoder is:
% ----------- equation-----------
\begin{align}
    & V_{mul} = [rt^{G^{one}_{sub}};rt^{G^{two}_{sub}}; E^{post}] \label{eq: multi-to-decoder}
\end{align}
% ----------- end of equation -----------
In the multi-hop scenario, Eq.~\ref{eq: multi-to-decoder} replaces Eq.~\ref{eq: one-to-decoder}. Empirically, we found the amount of two-hop knowledge is substantially larger than that of one-hop knowledge, and hence introduces noise and additional computational complexity. To address these issues, we choose the top 100 two-hop knowledge pieces that are most similar to the dialogue context based on sentence-transformer scores for our experiments.

% =============================== Section 4 ==================================
% \section{Experiments}
% \label{sec:experiments}

% ######################### Section 4.1 ########################
\section{Experimental Setup}
\label{sec:experimentalsetup}

\subsection{Datasets and Baselines} 
\noindent\textbf{Datasets.} Experiments are conducted on open-domain conversations extracted from Reddit \cite{zhou2018commonsense}. ConceptNet \cite{Conceptnet} is used as the commonsense knowledge base, which consists of $120,850$ triples, $21,471$ entities, and $44$ relations. The knowledge base contains not only world facts, but also common concepts. Each single-round conversation pair is preserved if it can be connected by at least one knowledge triple. The dataset has $3,384,185$/$10,000$/$20,000$ conversations for training/evaluation/testing, respectively.

% \subsubsection{Baselines}
% \label{sec:baselines}
\noindent\textbf{Baselines.} 
We compare our model against five competitive baselines used in this task. There are some similar works, e.g. \cite{yu-etal-2022-diversifying-content,wu2022improving}, which use external resources of documents or other kind of knowledge other than graph knowledge. They cannot be considered as our baseline models. Our research focuses on exploring a more efficacious approach for the integration of heterogeneous features within a language model framework. Consequently, large-scale language models, exemplified by ChatGPT~\footnote{ChatGPT, a recent language model release by \url{https://chat.openai.com/}, boasts a parameter count approximately 100 times greater than that of our base language model, UniLM.}, are neither employed as the primary language model in our experiments nor included within the baseline models under examination.

\begin{itemize}
[noitemsep,nolistsep,leftmargin=*]
    \item \textbf{Seq2seq} \cite{Seq2seq}: A widely used encoder-decoder in conversational systems.
    \item \textbf{MemNet} \cite{ghazvininejad2018knowledge}: A model which uses MemNet to store knowledge triples.
    \item \textbf{CopyNet} \cite{Copynet}: A model which copies concepts in knowledge triples to generate responses.
    \item \textbf{CCM} \cite{zhou2018commonsense}: The SOTA model for one-hop knowledge-enhanced dialogue which leverages two graph-attention mechanisms and CopyNet to model one-hop knowledge triples and incorporate knowledge concepts into responses.
    \item \textbf{ConceptFlow} \cite{zhang-etal-2020-grounded}: The SOTA model for multi-hop knowledge-enhanced dialogue which has a similar method to CCM but uses additional graph attention to model two-hop knowledge triples.
\end{itemize}

% Training and parameters details are given in Appendix \ref{apx:training}.

\subsection{Training Details and Parameters} 
%Refer to Appendix \ref{apx:training}
UniLM-base-cased is used as the pre-trained language model. It has $12$ BERT-block layers featuring $12$ attention heads in each layer. The first six layers of the model are considered to be an encoder and the last six layers a decoder. The word embedding size is $768$. The conversations and knowledge triples share the same BERT embedding layer, with a maximum length of $512$. The hidden representation of the sixth layer is used to facilitate the 2-layer knowledge aggregation model. An Adam optimizer is used with a batch size of 36. The learning rate is $5e^{-5}$. The model was trained on a \textit{Tesla V100} machine for approximately $7$ days, and $20$ epochs.

% ######################### Section 4.2 ########################
\subsection{Evaluation Protocol}
% \label{sec:evaluationmetrics}

\noindent\textbf{Automatic Evaluation Metrics.} We follow \cite{zhou2018commonsense} and \cite{DSTC7} in adopting the metrics of perplexity (PPL) \cite{Perplexity} and Entity Score (ES), and follow \cite{DSTC7} in adopting BLEU \cite{Bleu}, NIST \cite{Nist}, METEOR \cite{Meteor}, Dist, and Ent \cite{zhang2018generating}, where the Entity Score measures the average number of entities per response and others measure the quality of generated responses. BLEU, NIST, and METEOR are calculated between generated responses and golden responses, whilst Dist and Ent are calculated within generated responses.

\noindent\textbf{Human Evaluation.} Pair-wise comparisons are conducted with the most competitive baseline and the ablation model by five evaluators giving their preference of response on 100 randomly collected samples, regarding two aspects: the \textit{appropriateness} (whether the response is appropriate in the context) and \textit{informativeness} (whether the response contains new information).

% % % ######################### Section 4.3 ########################

\section{Experimental Results}

% ---------------------- Section 4.3.1 -------------------------
\subsection{Automatic Evaluation}
\label{sec:experimentresults}

\noindent\textbf{Referenced Metrics.} The experimental results shown in \autoref{reference-metrics} comprehensively measure the quality of the generated responses. It can be observed that 
our CADGE model (which uses one-hop knowledge) outperforms most of the baselines. For instance, it outperforms 
CCM, one of the SOTA models using one-hop knowledge, on all metrics,  obtaining at least twice the scores of the CCM (for BLEU-4, the difference is even almost four times). 
When compared to ConceptFlow, a SOTA model that exploits multi-hop knowledge, CADGE is still able to perform better (on over half of the metrics) or give comparable performance.

% ----------- tab: referenced metrics----------
\begin{table*}[t]
\centering \small
\resizebox{0.95\linewidth}{!}{
\begin{tabular}{l|cccc|cccc|c}
\toprule
\textbf{Model} & \textbf{BLEU-1} & \textbf{BLEU-2} & \textbf{BLEU-3} & \textbf{BLEU-4} & \textbf{NIST-1} & \textbf{NIST-2} & \textbf{NIST-3} & \textbf{NIST-4} & \textbf{METEOR} \\
\hline
\textbf{Seq2Seq} & 0.1702 & 0.0579 & 0.0226 & 0.0098 & 1.0230 & 1.0963 & 1.1056 & 1.1069 & 0.0611 \\
\textbf{MemNet} & 0.1741 & 0.0604 & 0.0246 & 0.0112 & 1.0975 & 1.1847 & 1.1960 & 1.1977 & 0.0632  \\
\textbf{CopyNet} & 0.1589 & 0.0549 & 0.0226 & 0.0106 & 0.9899 & 1.0664 & 1.0770 & 1.0788 & 0.0610  \\
\textbf{CCM} & 0.1413 & 0.0484 & 0.0192 & 0.0084 & 0.8362 & 0.9000 & 0.9082 & 0.9095 & 0.0630  \\
\textbf{ConceptFlow} & \textbf{0.2451} & \textbf{0.1047} & 0.0493 & 0.0246 & \uline{1.6137} & 1.7956 & 1.8265 & 1.8329 & \uline{0.0942} \\
\midrule
\textbf{CADGE} & 0.2078 & 0.0967 & \uline{0.0551} & \uline{0.0326} & 1.5566 & \uline{1.8113} & \uline{1.8609} & \uline{1.8683} & 0.0893 \\
\textbf{- w/o es-loss} & 0.2024 & 0.0937 & 0.0525 & 0.0315 & 1.5114 & 1.7421 & 1.7826 & 1.7878 & 0.0895 \\
\textbf{- w/o aggregation} & 0.1941 & 0.0920 & 0.0528 & 0.0322 & 1.4672 & 1.6994 & 1.7421 & 1.7477 & 0.0861 \\
\textbf{- w/o ca-gat} & 0.2019 & 0.0730 & 0.0305 & 0.0138 & 1.3562 & 1.4919 & 1.5082 & 1.5101 & 0.0796 \\
\midrule
\textbf{- w/ two hops} & \uline{0.2197} & \uline{0.1011} & \textbf{0.0558} & \textbf{0.0328} & \textbf{1.6689} & \textbf{1.9171} & \textbf{1.9606} & \textbf{1.9661} & \textbf{0.1053}\\
\bottomrule
\end{tabular}
}
\caption{\label{reference-metrics}
Automatic evaluation on popular reference-based metrics used in the task of open domain dialogue. The best performing model is highlighted in \textbf{bold}, and the second best is \uline{underlined}. \textbf{- w/o es-loss} denotes the ablated model without the auxiliary entity selection task; \textbf{- w/o aggregation} denotes the model without the feature aggregation process (which is implemented by directly mean pooling the features of flattened triples without our two layer forward aggregation process); \textbf{- w/o ca-gat} denotes the model without our proposed context-aware GAT introduced in \autoref{sec:contextawaregat}; \textbf{- w/ two hops} denotes the model expanded by two-hop knowledge introduced in \autoref{sec:multi-hop}.
}
\end{table*}
% ----------- the end of tab ----------

Given that the baselines contain the most representative framework for encoding heterogeneous features with separate encoders (i.e. CCM), the results clearly show the effectiveness of our knowledge aggregation mechanism, which better captures the heterogeneous features from the posts and knowledge facts with unified feature encoding and knowledge aggregation, and hence improves the quality of the generated responses. The ablation experiments further demonstrate the advances of the knowledge aggregation mechanism. Our context-aware GAT largely contributes to the improvement in performance, which can be observed in the comparison with \textit{- w/o ca-gat}. Additionally, we also tried to allow neural networks to understand the semantics by directly coagulating the features of flattened triples \textit{- w/o aggregation}, where the performance drops significantly, indicating the layer forward aggregation process is a key factor to the understanding of semantics contained in graph knowledge. By incorporating the enhanced two-hop knowledge, CADGE achieves universal performance gains on all metrics, further demonstrating the usefulness of incorporating multi-hot knowledge. 

% ----------- tab: unreferenced-----------
\begin{table}[t]
\centering \small
\resizebox{0.9\linewidth}{!}{
\begin{tabular}{l|cc|c}
\toprule
\textbf{Model} & \textbf{Dist-1} & \textbf{Dist-2} & \textbf{Ent-4}\\
\hline
\textbf{Seq2Seq}  & 0.0123 & 0.0525 & 7.665 \\
\textbf{MemNet}  & 0.0211 & 0.0931 & 8.418 \\
\textbf{CopyNet}  & 0.0223 & 0.0988 & 8.422 \\
\textbf{CCM}  & 0.0146 & 0.0643 & 7.847 \\
\midrule

\textbf{Conceptflow} & 0.0223 & 0.1228 & \uline{10.270} \\
\textbf{CADGE} & 0.0288 & 0.1136 & 10.141\\
\textbf{- w/d es-loss}  & 0.0326 & \uline{0.1242} & 9.445 \\
\textbf{- w/o aggregation} & \uline{0.0340} & 0.1234 & 8.968 \\
\textbf{- w/d ca-gat}  & 0.0189 & 0.0755 & 9.599 \\
\midrule
\textbf{- w/ two hops} & \textbf{0.0461} & \textbf{0.2702} & \textbf{11.626}\\
\bottomrule
\end{tabular}
}
\caption{\label{unreference-metrics}
Automatic evaluation on unreferenced metrics.}
\end{table}
% ----------- the end of tab ----------

\noindent\textbf{Unreferenced Metrics.}  
%To avoid the evaluation bias brought by human written samples,
We also examine the quality of the generated responses with unreferenced metrics that measure diversity and informativeness (entity score). As the results show in \autoref{unreference-metrics}, both language diversity and informativeness are substantially improved with our proposed knowledge aggregation framework. For example, the diversity score of our model is on par with that of the SOTA model (ConceptFlow). When two-hop knowledge is incorporated, the scores of CADGE are almost double that of ConceptFlow, which also uses multi-hop knowledge. 

These strong results demonstrate our model offers a substantial improvement over existing approaches when considering the language quality and relevance of generated responses, and matches better with the golden reference responses. When generating responses only with the UniLM model, performance on all metrics drops substantially, further demonstrating that the proposed Context-aware GAT contributes immensely to generating informative and high-quality responses via effective aggregation of knowledge triples. Both the referenced and unreferenced metrics indicate that with the improvement in heterogeneous feature capturing and global feature aggregation, CADGE can better exploit background knowledge to generate more high-quality and human-like responses.

% ----------- tab:entity score evaluation-----------
\begin{table*}[tb]
\centering
\resizebox{0.75\linewidth}{!}{
\begin{tabular}{l|c|c|c|c|c|c|c|c|c|c}
\toprule
\multirow{2}{*}{\textbf{Model}} & \multicolumn{2}{c|}{\textbf{Overall}} & \multicolumn{2}{c|}{\textbf{High Freq.}} & \multicolumn{2}{c|}{\textbf{Medium Freq.}} & \multicolumn{2}{c|}{\textbf{Low Freq.}} & \multicolumn{2}{c}{\textbf{OOV}}\\
% \cline{2-11}
 & PPL$\downarrow$ & ES$\uparrow$ & PPL$\downarrow$ & ES$\uparrow$ & PPL$\downarrow$ & ES$\uparrow$ & PPL$\downarrow$ & ES$\uparrow$ & PPL$\downarrow$ & ES$\uparrow$ \\
\midrule
\textbf{Seq2Se} & 47.02 & 0.72 & 42.41 & 0.71 & 47.25 & 0.74 & 48.61 & 0.72 & 49.96 & 0.67 \\

\textbf{MemNet} & 46.85 & 0.76 & 41.93 & 0.76 & 47.32 & 0.79 & 48.86 & 0.76 & 49.52 & 0.71 \\

\textbf{CopyNet} & 40.27 & 0.96 & 36.26 & 0.91 & 40.99 & 0.97 & 42.09 & 0.96 & 42.24 & 0.96 \\

\textbf{CCM} & 39.18 & 1.18 & 35.36 & 1.16 & 39.64 & 1.19 & 40.67 & 1.20 & 40.87 & 1.16 \\
\midrule
\textbf{CADGE} &  \textbf{33.99 }& \textbf{1.39} & \textbf{31.50} & \textbf{1.49} & \textbf{34.39} & \textbf{1.43} & \textbf{34.67} & \textbf{1.35} & \textbf{35.56} & \textbf{1.29} \\
\textbf{- w/o es-loss} & 34.73 & 1.28 & 32.31 & 1.36 & 35.18 & 1.33 & 35.41 & 1.24 & 35.19 & 1.19 \\
\textbf{- w/o aggregation} & 34.71 & 1.35 & 32.25 & 1.42 & 35.16 & 1.39 & 35.36 & 1.31 & 35.62 & 1.27\\
\textbf{- w/o ca-gat} & 36.51 & 1.03 & 33.82 & 1.10 & 37.02 & 1.06 & 37.23 & 1.01 & 38.12 & 0.95 \\
\bottomrule
\end{tabular}
}
\caption{\label{ppl and ent score}
Automatic evaluation on the metrics of \textit{perplexity} ($\downarrow$) and \textit{entity score} ($\uparrow$). The experiment is set up with one-hop knowledge. Therefore ConceptFlow, which needs two-hop knowledge, is excluded in this experiment. The test set (\textbf{Overall}) is categorised into 4 sub-datasets with different frequencies (\textbf{Freq.} and \textbf{OOV} (out of vocabulary)) of the entities included in the posts. The overall PPL and ES of \textbf{ConceptFlow} are 36.51 and 1.03, respectively. The overall PPL and ES of Cadge \textbf{- w/ two hops} are 29.90 and 1.68, respectively. Since ConceptFlow did not evaluate frequency grouped test data on two-hop data, we only compare models with one-hop data here.
}
\end{table*}
% ----------- the end of tab -----------

% ---------------------- Section 4.3.2 -------------------------
\subsection{Analysis of the Knowledge Aggregation Mechanism}

\noindent\textbf{Perplexity and Entity Score.} 
Based on the frequency of words in the posts, we divide the test set into four sections (high, middle, low, and OOV) in order to evaluate the performance and robustness of each model when faced with frequently seen dialogues as well as uncommon dialogues. For a fair comparison, we limit the retrieved knowledge to one-hop as not every baseline is able to incorporate multi-hop knowledge (e.g., CCM). 
As shown in \autoref{ppl and ent score},  our model achieves the lowest perplexity and the highest entity scores for all frequency groups. The lowest perplexity indicates that the proposed model achieves the best predictive performance of the language model and generates a more fluent response than other baselines, while the best entity scores indicate that the proposed model better exploits graph features to select appropriate entities contained in the post. For the ablation study, we compare CADGE to the base model UniLM,\footnote{The ablated model \textbf{- w/o ca-gat} is regarded as the base model UniLM, which works without graph knowledge.} which is a pre-trained language model without the Context-aware GAT. The substantial performance gain of CADGE over UniLM  demonstrates the importance of leveraging global features obtained by graph knowledge to improve both the model's understanding and generation ability. 

% ----------- fig: attention distribution -----------
\begin{figure}[tb]
\centering
\includegraphics[width=0.99\linewidth]{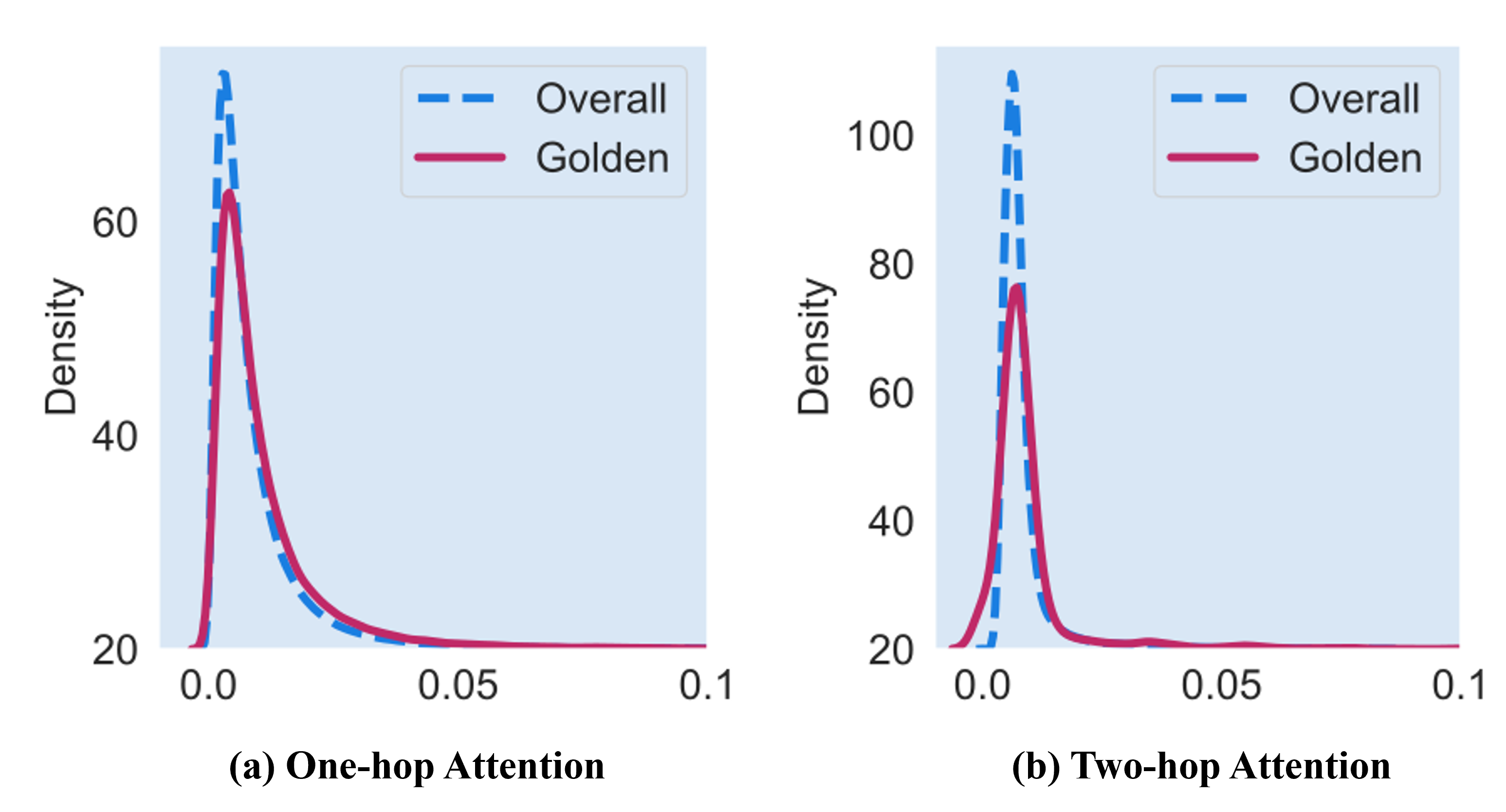}
\caption{The learned attention probability density curves on knowledge facts.}
\label{one-hop-attn}\label{two-hops-attn}
\end{figure}
% ----------- the end of fig -----------

\noindent\textbf{Attention Distribution on Knowledge.} In order to test whether our model has learned to place more attention on golden knowledge facts for dialogue generation, we draw probability density curves to compare the attention distribution of golden knowledge (i.e. retrieved knowledge facts that appear in \textit{reference responses}) and overall knowledge (knowledge facts retrieved from  posts). Figure \ref{one-hop-attn} illustrates the result with one-hop knowledge aggregation, and Figure \ref{two-hops-attn} with two-hop. It can be observed that Context-aware GAT is able to learn to select more related knowledge facts for dialogue generation, as demonstrated by the curves showing that golden knowledge facts have a higher probability of having higher attention scores. In other words, our graph model is able to obtain an aggregated representation that places more focus on relevant knowledge for response generation.

% ----------- fig: attn box figure -----------
\begin{figure}[t]
\centering
\includegraphics[width=0.75\columnwidth]{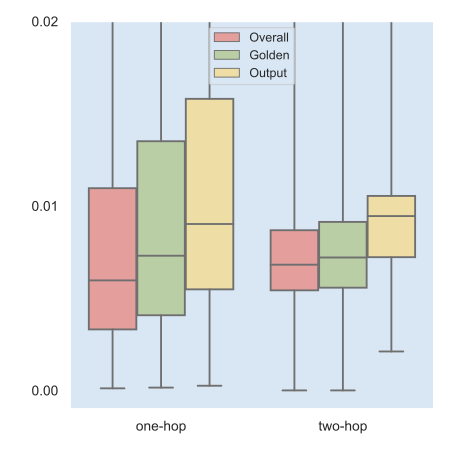}
\caption{A box plot to analyse attention scores learned by context-aware GAT to aggregate features from one-hop and two-hop knowledge. \textbf{Overall}: average attention of all knowledge; \textbf{Golden}: average attention of golden knowledge; \textbf{Output}: average attention of knowledge in generated responses. }
\label{fig:attention comparing}
\end{figure}
% ----------- end of fig -----------

% ----------- tab: human evaluation ----------
\begin{table*}[t]
\centering 
\resizebox{0.95\linewidth}{!}{
\begin{tabular}{r|lc|c|lc|c|lc|c}
\toprule[1pt]
\multirow{2}{*}{\textbf{Choice \%}} & \multicolumn{3}{c}{\textbf{CADGE$_{one\_hop}$ \textit{vs} CCM}} & \multicolumn{3}{c}{\textbf{CADGE$_{one\_hop}$ \textit{vs} \textbf{\textit{- w/o ca-gat}}}} & \multicolumn{3}{c}{\textbf{CADGE$_{two\_hops}$ \textit{vs} ConceptFlow}} \\
\cline{2-10} 
& \textbf{CADGE$_{one\_hop}$} & \textbf{CCM} & \textbf{$\mathit{Kappa}$} & \textbf{CADGE$_{one\_hop}$} & \textbf{\textit{- w/o ca-gat}} & \textbf{$\mathit{Kappa}$} & \textbf{CADGE$_{two\_hops}$} & \textbf{Conceptflow} & \textbf{$\mathit{Kappa}$}  \\
\midrule
\textbf{\textit{App.}} & \textbf{66.0} & 34.0 & 0.367  & \textbf{58.1} & 41.9 & 0.323   & \textbf{64.7} & 35.3 & 0.321  \\
\textbf{\textit{Inf.}} & \textbf{63.3} & 36.7 & 0.278  & \textbf{60.1} & 39.9 & 0.318  & \textbf{64.9} & 35.1 & 0.304  \\

\bottomrule
\end{tabular}
}
\caption{\label{human evaluation}
Human Evaluation w.r.t.~\textit{appropriateness} and \textit{informativeness}. The score is the percentage that the proposed model wins against its competitor. $\mathit{Kappa}$ denotes Fleiss’ Kappa~\cite{fleiss1971measuring}, which indicates all of our evaluation annotations reach a fair agreement. The proposed model is significantly better (sign test, $p < 0.005$).
}
\end{table*}

% ----------- end of tab ----------

\noindent\textbf{Statistics for Attention Scores.} To better analyse the statistics of the learned attention scores during the knowledge aggregation in our model, we further draw a box plot to compare the attention scores of different knowledge facts, with the results shown in \autoref{fig:attention comparing}.  According to the attention scores distribution, the knowledge facts in the output have higher attention than other retrieved knowledge, meaning the model has more confidence to select related knowledge to generate responses.\footnote{If the generated knowledge facts have the same distribution as the overall, this means that the model is confused when selecting relevant knowledge facts.} With respect to the attention on the golden knowledge facts, they are substantially different from other retrieved knowledge, which demonstrates that with the knowledge aggregation process, our framework learned the correct features to represent knowledge facts, leading to more appropriate selections over retrieved knowledge facts.

% ---------------------- Section 4.3.3 -------------------------
\subsection{Human Evaluation}

We also conducted human evaluation to further consolidate our model performance. The results are presented in \autoref{human evaluation}, which, in accordance with the previously presented automatic metrics, demonstrates that our model outperforms the SOTA baselines on both \textit{appropriateness} and \textit{informativeness}, and proves the effectiveness of the proposed Context-Aware GAT. Under the condition of either one-hop knowledge or two-hop knowledge, 
CADGE achieves significant improvements in producing more informative and appropriate responses, owing to the proposed context-aware knowledge aggregation framework. 
% ----- fig: Visualisation of the knowledge aggregation process -----------
\begin{figure*}[tb]
\centering
\includegraphics[width=0.85\linewidth]{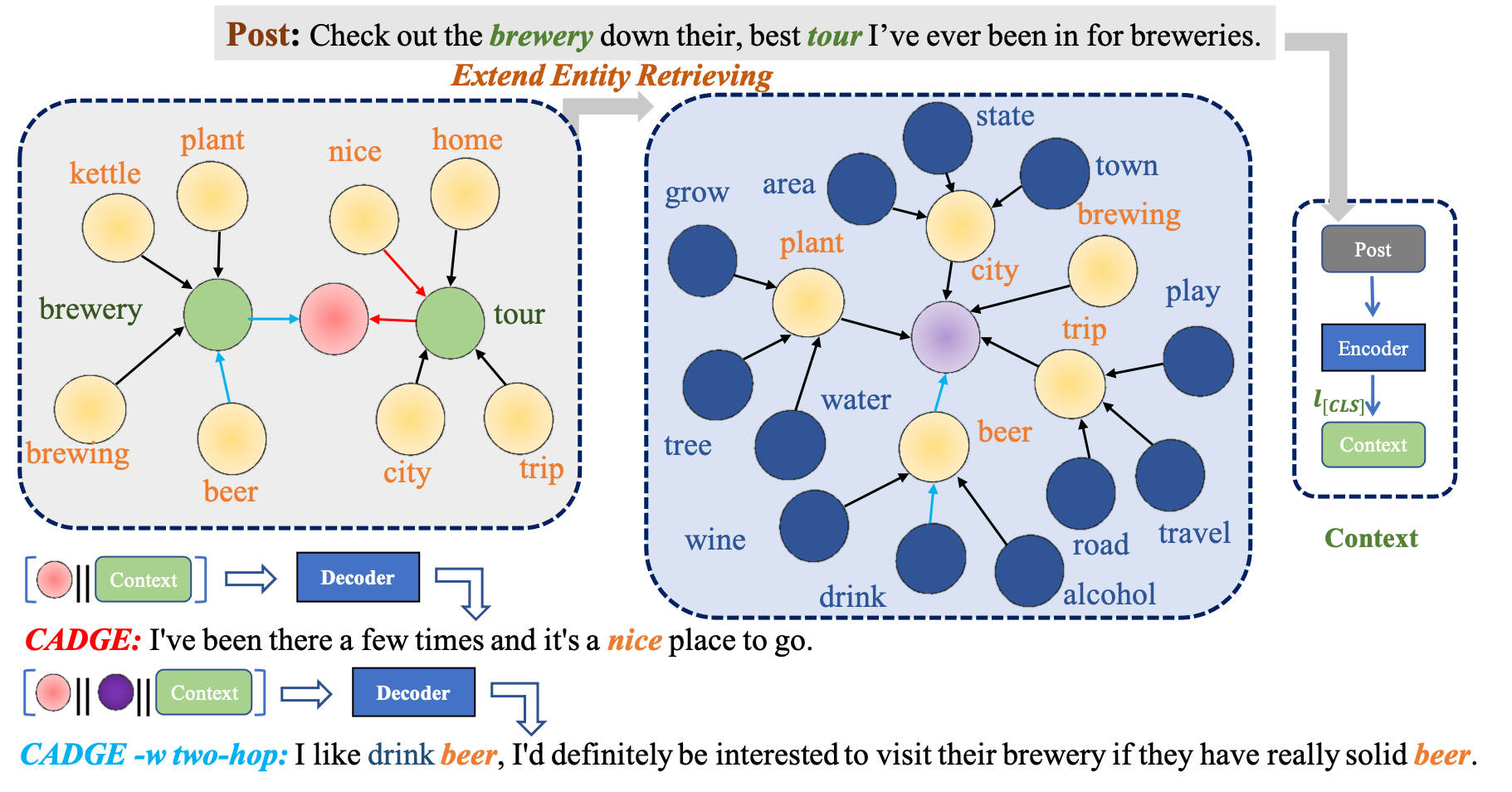}
\caption{Visualization of the knowledge aggregation process with an example.}
\label{fig:visualisation}
\end{figure*}
% ----------- end of fig -----------

\subsection{Knowledge Aggregation Process.}

In \autoref{fig:visualisation}, we illustrate an example of the knowledge aggregation process of our framework, where the left subgraph represents the one-hop knowledge aggregation (i.e. yellow nodes) and the right subgraph represents the additional knowledge aggregation attending to the second-hop knowledge (i.e. blue nodes). As mentioned in \S\ref{sec:methodology}, CADGE aggregates features layer by layer. For one-hop CADGE, the aggregated representation (the red node) of all retrieved knowledge facts is concatenated with the context features of the post, and fed into the neural decoder to generate responses. When incorporating two-hop knowledge, CADGE exploits a similar mechanism, and we obtain an additional knowledge representation (the purple node) for response generation. It can be seen from the example that when CADGE only uses one-hop knowledge, it selects ``nice'' from the graph which is subsequently used to generate a response. When two-hop knowledge is available,  CADGE selects ``beer'' from the one-hop graph and ``drink'' from the two-hop graph, improving informativeness and making the response more interesting. 
We also provide a detailed qualitative analysis of sample responses from the one-hop and two-hop knowledge experiments in Appendix \ref{apx:case-study}.

% =============================== Section 4 ==================================
\section{Conclusion}
\label{sec:conclusion}
In this paper, we proposed a novel knowledge aggregation framework for the knowledge graph enhanced dialogue generation task. This framework implements a Context-aware GAT which applies representation learning of the heterogeneous features from graph knowledge text, and the neural networks effectively learn to incorporate globally aggregated features to enhance response generation with rich representations. Extensive experiments are conducted to demonstrate that our framework outperforms SOTA baselines on both automatic and human evaluation, as the proposed Context-Aware GAT largely improved the semantic understanding of both graph and text knowledge to enhance the appropriateness and informativeness of generated responses. The expansion of Context-Aware GAT to two-hop knowledge also indicates the robustness and effectiveness of our framework in increasing the amount of grounded graph knowledge in responses. We hope that our proposed framework can benefit research in all text generation tasks where knowledge graphs are incorporated, and transferable research will be continued in further work.

% =============================== Acknowledgements ==================================

\section*{Acknowledgements}
Chen Tang is supported by the China Scholarship Council (CSC) for his doctoral study (File No.202006120039). Tyler Loakman is supported by the Centre for Doctoral Training in Speech and Language Technologies (SLT) and their Applications funded by UK Research and Innovation [grant number EP/S023062/1]. 
%We also gratefully acknowledge the anonymous reviewers for their insightful comments.

% =============================== Ethics Statement ==================================
\section*{Ethics Statement}
All work presented within this paper is in line with the ethical code of conduct of both ACL and the institutions of the authors. In this work we present a method to increase the level of world knowledge in dialogue system responses. In turn, this results in the applications being more useful to the end user, and being better positioned to answer a range of topics. However, we acknowledge that using similar approaches to incorporate knowledge into dialogue systems should be cautious of the veracity and validity of the utilised "knowledge" in order to avoid issues relating to misinformation. We do not motivate this work in terms of a specific application, and instead present a method for incorporating knowledge graph structured information in general.

\normalem
\bibliography{custom}

\appendix

% ====================
\section{Related Work} \label{apx:related-works}
%\subsection{Open-domain Dialogue Generation}
%Large language models such as UniLM \cite{dong2019unified}, GPT-2 \cite{radford2019language}, and BART \cite{bart} are widely used in open domain dialogue generation systems~\cite{zeng-etal-2021-affective}. DialoGPT \cite{diagpt} was pre-trained on a dialogue dataset containing 147M conversations and is based on the autoregressive GPT-2 model, using a maximum mutual information (MMI) scoring function to address the low amount of information in the generated text. \cite{meena} built a 2.6B-parameter Evolved Transformer architecture to model the relation between context-response pairs. To generate more informative responses, \cite{bao2019plato,bao2020plato} use latent variables to model one-to-many relationships in context-response pairs. Finally, \cite{roller-etal-2021-recipes} use a retrieval model to retrieve candidate responses and then concatenates them to represent the context before inputting them into the transformer to generate the model.

\noindent\textbf{Knowledge Enhanced Dialogue Generation.}  
Recently, much work has focused on augmenting dialogue systems with additional background knowledge. Such works can be divided into dialogue systems augmented with unstructured knowledge, and those augmented with structured knowledge. With unstructured knowledge, \cite{komeili-etal-2022-internet} models web page information and feeds it into a language model. \cite{ghazvininejad2018knowledge} and \cite{lotfi-etal-2021-teach} encode the filtered factual statements with a specific encoder and then pass them into the decoder along with context. \cite{lian2019response} use context to aggregate knowledge statements and find that aggregated knowledge gives better results than filtered knowledge. Regarding structured knowledge, graph neural networks \cite{GNN} are usually used to embed graph information to input into a language model. \cite{zhou2018commonsense} uses GRUs and two graph attention modules to select appropriate triples to incorporate into responses. In order to exploit the benefits of multi-hop knowledge, \cite{zhang-etal-2020-grounded} adds an attention mechanism in a similar way to filter the appropriate knowledge. Finally, \cite{tuan2019dykgchat} proposes a model which selects the output from a sequence-to-sequence model and a multi-hop reasoning model at each time step.

\begin{table}[t]
\centering
\resizebox{\columnwidth}{!}{
\begin{tabular}{l|l}
\toprule[2pt]
\textbf{\#1 Post} & \textbf{\makecell[l]{So if op has ever taken a road trip he was at more of a risk \\ than the surgery he's\\ gonna have :d}}\\
\midrule[1pt]
CADGE$_{one\_hop}$ & \makecell[l]{I'm not saying it's \textcolor{red}{safe}, but it's not like he's going to \\be in a \textcolor{red}{car} accident.} \\
\midrule
CCM & I'm not sure if he 's a \textcolor{red}{surgeon} or not . \\
\midrule
\textit{- w/o ca-gat} & \makecell[l]{I'm not saying he's not going to be taking a road trip.} \\ 
\bottomrule[1pt]

\textbf{\#2 Post} & \textbf{\makecell[l]{Because i haven't heard it on the radio enough.}} \\ 
\midrule[1pt]
CADGE$_{one\_hop}$ & \makecell[l]{I've heard it on radio. it's a pretty good \textcolor{red}{song}} \\
\midrule
CCM & \makecell[l]{I'm not sure if it's a radio station or not. \\i 'm not sure if it 's a radio station or not.}  \\ 
\midrule
\textit{- w/o ca-gat} & \makecell[l]{I've heard it on the radio.}  \\
\bottomrule[1pt]

\textbf{\#3 Post} & \textbf{\makecell[l]{We will have to agree or disagree mate!}} \\ 
\midrule[1pt]
CADGE$_{one\_hop}$ & \makecell[l]{I'm not \textcolor{red}{disagreeing}, i'm just saying that it's not a valid \textcolor{red}{argument}.} \\
\midrule
CCM & \makecell[l]{I'm not saying it 's a bad \textcolor{red}{argument} , \\but i don't think it 's a good idea to be a fan of \\ the game.}  \\ 
\midrule
\textit{- w/o ca-gat} & \makecell[l]{I'm not your mate, pal.}  \\
\bottomrule[1pt]

\textbf{\#4 Post} & \textbf{\makecell[l]{What do you mean by the auto dude?}} \\ 
\midrule[1pt]
CADGE$_{one\_hop}$ & \makecell[l]{I mean the \textcolor{red}{guy} who's in the \textcolor{red}{car}.} \\
\midrule
CCM & \makecell[l]{I'm not sure . i 'm not sure if i 'm wrong . i 'm just saying that \\ the \textcolor{red}{guy} is a \textcolor{red}{guy} who  doesn't know what \\he 's talking about .}  \\ 
\midrule
\textit{- w/o ca-gat} & \makecell[l]{Auto \textcolor{red}{man}.}  \\

\bottomrule[2pt]
\end{tabular}
}
\caption{\label{case study one}
Sample responses from the one-hop knowledge experiments, comparing CADGE with the SOTA CCM, and its ablation model UniLM. Tokens in \textcolor{red}{red} are entities in knowledge triples.
}

\end{table}
% ----------- the end of tab ----------

% ----------- tab: case study of two hops ----------
\begin{table}[t]
\centering
\resizebox{\columnwidth}{!}{
\begin{tabular}{l|l}
\toprule[2pt]
\textbf{\#1 Post} & \textbf{\makecell[l]{My brain must not like music, i've never gotten \\chills from a song.}} \\ 
\midrule[1pt]
CADGE$_{two\_hop}$ & \makecell[l]{I don't really \textcolor{red}{listen} to music at all but i have found some \\ \textcolor{red}{songs} that you may \textcolor{blue}{enjoy} like.}  \\
\midrule
ConceptFlow & \makecell[l]{The whole \textcolor{red}{album} is a \textcolor{red}{great} song.} \\
\midrule
CADGE$_{one\_hop}$ & \makecell[l]{I don't even \textcolor{red}{listen} to music, I don't \textcolor{red}{love} \\ music and i have no idea how to get chills from something .}  \\ 

\bottomrule[1pt]
\textbf{\#2 Post} & \textbf{\makecell[l]{The rest of the series will be released for free as well.}} \\ 
\midrule[1pt]
CADGE$_{two\_hop}$ & \makecell[l]{I don't know about the rest of the \textcolor{red}{collection}, i only really \textcolor{blue}{read} \\ the second \textcolor{blue}{book} .}  \\
\midrule
ConceptFlow & \makecell[l]{The whole \textcolor{red}{album} is a \textcolor{red}{great} song.} \\
\midrule
CADGE$_{one\_hop}$ & \makecell[l]{I don't really \textcolor{red}{care} if it's on sale, i can just \textcolor{red}{play} it on steam.}  \\ 

\bottomrule[1pt]
\textbf{\#3 Post} & \textbf{\makecell[l]{Check out the brewery down their, best tour ive \\ ever been in for breweries}} \\ 
\midrule[1pt]
CADGE$_{two\_hop}$ & \makecell[l]{I like \textcolor{blue}{drink} \textcolor{red}{beer}, i'd definitely be interested to visit their \\brewery if they have really solid beer.}  \\
\midrule
ConceptFlow & \makecell[l]{I've been to the brewery. it's a blast to see.} \\
\midrule
CADGE$_{one\_hop}$ & \makecell[l]{i've been there a few times and it's a \textcolor{red}{nice} place to go.}  \\ 

\bottomrule[1pt]
\textbf{\#4 Post} & \textbf{\makecell[l]{I'm thoroughly disappointed that i don't have work - \\appropriate white clothing for today.}} \\ 
\midrule[1pt]
CADGE$_{two\_hop}$ & \makecell[l]{I just go through some tweet to find a better \textcolor{blue}{outfit} to \textcolor{red}{wear}.}  \\
\midrule
ConceptFlow & \makecell[l]{I'll have to \textcolor{blue}{check} out my new white \textcolor{red}{shirts} for the next day} \\
\midrule
CADGE$_{one\_hop}$ & \makecell[l]{You have to wear a white \textcolor{red}{shirt} to work for a few days a week.}  \\ 

\bottomrule[2pt]
\end{tabular}
}
\caption{\label{case study two}
Sample responses from the two-hop knowledge experiments, comparing CADGE$_{two\_hop}$ with the SOTA model ConceptFlow, and its one-hop ablation model CADGE$_{one\_hop}$. Tokens in \textcolor{red}{red} are entities belonging to the one hop knowledge, while tokens in \textcolor{blue}{blue} belong to the second hop.
}

\end{table}
% ----------- the end of tab ----------

% =============================
\section{Case Study} \label{apx:case-study}
Several post-response pairs are presented to compare generated responses by the proposed CADGE, the ablated model UniLM, and the SOTA models of CCM and ConceptFlow, in different external knowledge settings (one-hop or two-hop). \autoref{case study one} presents comparisons based on one-hop knowledge. Considering the ablation of external knowledge, it can be observed that without the representations for the knowledge facts, the UniLM model struggled to understand the context semantics and provide informative responses. For example, in the first post, the one-hop CADGE understands that the focus of the post is on "risk", while UniLM considers it to be on "road trip". In the third post, the one-hop CADGE understands that the focus of the post is on "agree", while UniLM considers it to be "mate".

When we consider the effectiveness of knowledge fact exploitation, the difference can be observed in generated responses between the one-hop CADGE and the CCM model. Responses from CADGE appear to be more logical and fluent than CCM. For instance, in the fourth post, the one-hop CADGE understands the phrase "auto dude" and gives an accurate explanation, instead of saying "not sure" as CCM does. The same phenomenon also appears in the first and second posts, which demonstrates that with the proposed knowledge aggregation framework, CADGE is more able to understand knowledge facts, and provide more informative and appropriate answers with this knowledge. 

In regards to the expansion on two-hop knowledge, our context GAT sustains the effectiveness and efficiency of knowledge representation learning. The additional comparisons are compared among CADGE$_{one\_hop}$, CADGE$_{two\_hop}$, and ConceptFlow in \autoref{case study two}. It can be observed that when the knowledge amount increases, CADGE$_{two\_hop}$ is better able to consider background knowledge when generating responses. For example, in the second and third post, CADGE$_{two\_hop}$ considers more retrieved knowledge facts to generate a response which results in responses with better quality, and that are more informative. In addition, the extra knowledge also gives more context semantics leading to better understanding of the dialogues. For instance, in all of the aforementioned cases, compared to one-hop CADGE and ConceptFlow, the two-hop CADGE chooses more informative concepts from all available knowledge, making the generated responses more interesting.
\end{document}